%% file: main.tex
\documentclass[10pt,twocolumn,letterpaper]{article}

\usepackage[final]{cvpr}      

\input{preamble}

\definecolor{cvprblue}{rgb}{0.21,0.49,0.74}
\usepackage[pagebackref,breaklinks,colorlinks,allcolors=cvprblue]{hyperref}
\usepackage{capt-of}
\usepackage{cuted} 
\def\confName{CVPR}
\def\confYear{2025}
\title{SHaDe: Compact and Consistent Dynamic 3D Reconstruction via Tri-Plane Deformation and Latent Diffusion}

\author{Asrar Alruwayqi\\
The Robotics Institute, Carnegie Mellon University \\ 
{\tt\small aalrwiqi@alumni.cmu.edu}
}

\begin{document}
\maketitle
\begin{strip}
  \centering
  \includegraphics[width=0.71\textwidth]{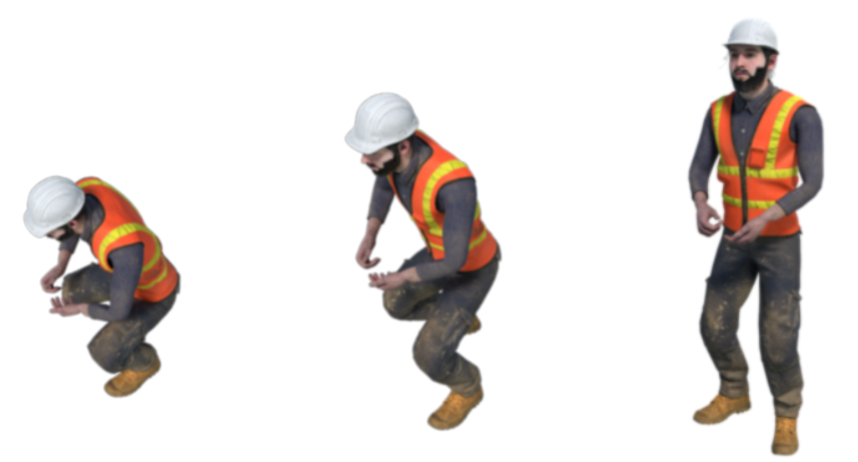}
  \vspace{1mm}
  
  \begin{minipage}{0.71\textwidth}
    \centering
    \small
    \hspace{0.11\linewidth} t = 0 \hspace{0.22\linewidth} t = 0.5 
    \hspace{0.22\linewidth} t = 1.0 (Canonical)
  \end{minipage}

  \captionof{figure}{
    \textbf{Visualization of our 4D scene reconstruction across time.}
    From left to right, we show reconstructed frames of a dynamic subject at increasing motion timestamps. 
    The rightmost frame corresponds to the canonical configuration into which all motion is explicitly warped for consistent appearance modeling.
    Our method preserves structural integrity and appearance fidelity over time, even under significant non-rigid deformation.
  }
  \label{fig:temporal_reconstruction}
\end{strip}

  


\input{sec/0_abstract}    
\input{sec/1_intro}
\input{sec/2_formatting}

\input{sec/3_finalcopy}
{
    \small
    \bibliographystyle{ieeenat_fullname}
    \bibliography{main}
}


\end{document}

%% file: preamble.tex
\usepackage[ruled,vlined]{algorithm2e}
\usepackage{amsmath}
\usepackage{graphicx}
\usepackage{pgfplots}
\pgfplotsset{compat=1.18}
\usepackage{pgfplotstable}
\usepackage{booktabs}



%% file: sec/0_abstract.tex
\begin{abstract}
We present a novel framework for dynamic 3D scene reconstruction that integrates three key components: an explicit tri-plane deformation field, a view-conditioned canonical radiance field with spherical harmonics (SH) attention, and a temporally-aware latent diffusion prior. Our method encodes 4D scenes using three orthogonal 2D feature planes that evolve over time, enabling efficient and compact spatiotemporal representation. These features are explicitly warped into a canonical space via a deformation offset field, eliminating the need for MLP-based motion modeling.

In canonical space, we replace traditional MLP decoders with a structured SH-based rendering head that synthesizes view-dependent color via attention over learned frequency bands improving both interpretability and rendering efficiency. To further enhance fidelity and temporal consistency, we introduce a transformer-guided latent diffusion module that refines the tri-plane and deformation features in a compressed latent space. This generative module denoises scene representations under ambiguous or out-of-distribution (OOD) motion, improving generalization.

Our model is trained in two stages: the diffusion module is first pre-trained independently, and then fine-tuned jointly with the full pipeline using a combination of image reconstruction, diffusion denoising, and temporal consistency losses. We demonstrate state-of-the-art results on synthetic benchmarks, surpassing recent methods such as HexPlane and 4D Gaussian Splatting in visual quality, temporal coherence, and robustness to sparse-view dynamic inputs.
\end{abstract}

%% file: sec/1_intro.tex
\section{Introduction}
\label{sec:intro}

Reconstructing dynamic 3D scenes (often framed as 4D reconstruction) is a foundational challenge in computer vision with applications in AR/VR, robotics, and digital twins. While neural radiance fields (NeRF)~\cite{mildenhall2020nerf} and explicit factorized representations like EG3D~\cite{chan2021eg3d} have achieved high-quality static reconstruction, dynamic content introduces unique challenges: fast and non-rigid motion, entangled view-time dependencies, and vulnerability to out-of-distribution (OOD) motion. Moreover, many dynamic NeRF extensions rely on deep deformation MLPs and dense 3D structures, resulting in heavy memory usage and slow inference limiting their practicality for long sequences or real-time applications.

This work addresses these limitations through three core innovations: (1) a fully explicit tri-plane deformation field, (2) a canonical radiance field based on spherical harmonics (SH) with dynamic attention, and (3) a temporally-aware latent diffusion prior. Our architecture enables efficient, temporally consistent reconstruction of dynamic 3D scenes from sparse multi-view imagery. Scene features are encoded using three orthogonal 2D planes (\(\mathcal{F}_{xy}, \mathcal{F}_{yz}, \mathcal{F}_{xz}\)) conditioned on time \(t\), avoiding dense 3D computation while capturing spatiotemporal context.

The deformation field is directly stored on tri-planes, without using MLPs, inspired by voxel-based methods such as HexPlane~\cite{HexPlane}. Query features are warped into a canonical space, where a hybrid radiance field predicts density from tri-plane features and synthesizes color through a novel dynamic SH attention mechanism. This replaces heavy MLPs for view-dependent modeling with a lightweight and interpretable SH decoder modulated by view direction and time.

To ensure robustness to fast motion, occlusions, and ambiguous observations, we introduce a latent diffusion model that refines the compressed tri-plane representations. A transformer-based encoder projects tri-plane tokens into a latent space, and a denoising diffusion model, trained using a DDPM-style~\cite{ho2020denoising}  objective and sampled via DDIM~\cite{ddim}, learns a prior over plausible dynamic scene evolution.

Our method emphasizes both efficiency and scalability. Plane-based factorization and SH-based decoding reduce memory and computation overhead compared to traditional volumetric MLPs. Meanwhile, latent diffusion operates in a compact space, enabling scalable training over long dynamic sequences. We demonstrate that our framework consistently outperforms recent state-of-the-art methods, including HexPlane~\cite{HexPlane} and 4D Gaussian Splatting (4D-GS)~\cite{4dgs}, in terms of reconstruction fidelity.
project page: https://asrarh.github.io/shade-project-page

%% file: sec/2_formatting.tex
\section{Related Work}
\label{sec:related}

\paragraph{Classical Reconstruction and Early View Synthesis.}
Traditional 3D reconstruction methods such as Structure-from-Motion (SfM) and Multi-View Stereo (MVS)~\cite{hartley2003multiple, seitz2006photo} rely on explicit geometry and camera calibration but often fail under non-rigid motion or sparse-view conditions. Early learning-based view synthesis~\cite{levoy1996light} improved photorealism but lacked temporal coherence and 3D consistency in dynamic scenarios.

\paragraph{Neural Scene Representations.}
Neural Radiance Fields (NeRF)~\cite{mildenhall2020nerf} introduced continuous volumetric rendering for static scenes. Follow-ups like NeRF-W~\cite{martinbrualla2020nerfw}, SRNs~\cite{sitzmann2019scene}, DeepSDF~\cite{park2019deepsdf}, and Occupancy Networks~\cite{mescheder2019occupancy} explored compact, implicit 3D encodings, though they remained focused on static or rigid content.

\paragraph{Dynamic Neural Fields.}
Modeling scene deformation over time led to dynamic NeRF variants such as D-NeRF~\cite{dnerf}, NeRFies~\cite{nerfies}, and HyperNeRF~\cite{park2023hypernerf}, which use MLP-based deformation fields. Later methods like TiNeuVox~\cite{TiNeuVox} improved speed and temporal modeling but remain computationally intensive and less robust to out-of-distribution (OOD) motion.

\paragraph{Plane-Based Representations.}
Tri-plane decomposition has emerged as a compact and structured alternative to dense 3D grids. EG3D~\cite{chan2021eg3d} pioneered tri-plane representations for generative 3D modeling. HexPlane~\cite{HexPlane} and K-Planes~\cite{kplanes} extended this idea to dynamic scenes by storing temporal and appearance-aware features across multiple planes. Our method builds on this paradigm by encoding deformation explicitly in tri-planes—without relying on MLPs improving both interpretability and inference speed.

\paragraph{Spherical Harmonics for Radiance Decoding.}
Spherical Harmonics (SH) are commonly used for efficient radiance prediction~\cite{Plenoxels}, typically in static voxel grids. Planeoxel~\cite{Plenoxels} demonstrated SH coefficient storage on orthogonal planes, but without dynamic modulation. We propose a novel SH attention decoder conditioned on time and viewing direction, enabling dynamic appearance modeling in a compact representation.

\paragraph{Diffusion Models for 3D Learning.}
Denoising Diffusion Probabilistic Models (DDPM)~\cite{ho2020denoising} and Latent Diffusion Models (LDM)~\cite{rombach2022high} have been adapted to 3D settings in works like DreamFusion~\cite{dreamfusion}, Magic3D~\cite{magic3d}, and Score Jacobian Chaining~\cite{sjc}. These methods typically distill gradients from pretrained 2D diffusion models into 3D fields. Recent advances such as ScoreHMR~\cite{stathopoulos2024score} and Point-Diffusion~\cite{pointcloudpretrainingdiffusion} extend diffusion to meshes and point clouds. Unlike these approaches, our model directly integrates latent diffusion into dynamic 3D reconstruction using a transformer-based encoder, refining the tri-plane and deformation features during training and inference.

\paragraph{Point-Based Rendering and 4D Gaussian Splatting.}
Gaussian Splatting (GS)~\cite{3dgs} enables high-quality, real-time rendering from dense-view video using point-based volumetric primitives. Extensions like 4D Gaussian Splatting~\cite{4dgs} target dynamic scenes but depend on dense multi-view capture and are less effective under sparse-view or canonical-space settings. In contrast, our method targets sparse input and compact latent refinement, offering an efficient alternative for 4D scene modeling.


\paragraph{Summary.}
While prior works explore deformation modeling~\cite{park2023hypernerf}, plane-based encoding~\cite{chan2021eg3d, HexPlane, kplanes}, spherical decoding~\cite{Plenoxels}, and diffusion priors~\cite{dreamfusion, rombach2022high}, our method presents a unified pipeline that:
(1) replaces deformation MLPs with explicit tri-planes,
(2) introduces a novel SH-attention field for radiance synthesis, and
(3) incorporates a transformer-based diffusion model for latent refinement enabling efficient, coherent 4D reconstruction from sparse views.

%% file: sec/3_finalcopy.tex
\section{Method}
\label{sec:method}

\begin{figure*}[t]
  \centering
  \includegraphics[width=\textwidth]{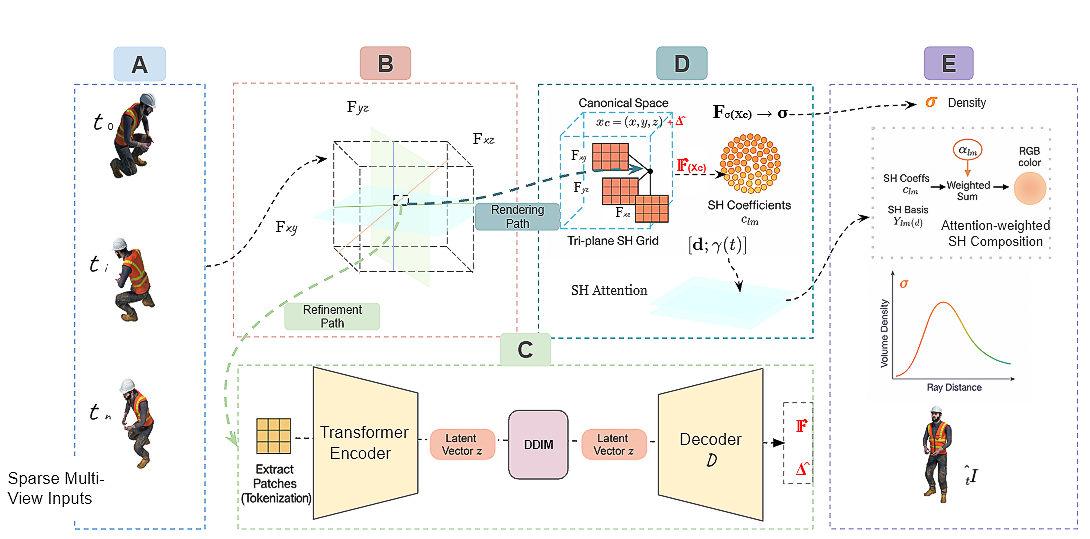}
  \caption{Overview of our dynamic scene reconstruction framework. 
  \textbf{(A)} Sparse multi-view inputs at different timesteps are provided as input. 
  \textbf{(B)} A tri-plane feature volume encodes spatial and temporal information across three orthogonal planes (\(F_{xy}, F_{yz}, F_{xz}\)). 
  \textbf{(C)} These features are tokenized and passed through a transformer encoder, producing a latent vector \(\mathbf{z}\) refined via a latent diffusion model (Refinement Path). The decoder reconstructs enhanced tri-plane features \(\hat{\mathcal{F}}\) and deformation offsets \(\hat{\Delta}\). 
  \textbf{(D)} In parallel (Rendering Path), the original tri-plane features are used to compute a deformation offset \(\Delta \mathbf{x}\), which warps query points into canonical space. SH coefficients are retrieved, and attention weights \(\alpha_{lm}(\mathbf{d}, t)\) are applied over SH basis functions \(Y_{lm}(\mathbf{d})\). 
  \textbf{(E)} The view- and time-aware SH composition yields color, while volume density \(\sigma\) is predicted from a separate tri-plane. These outputs are used for differentiable volume rendering to produce the final photorealistic output \(\hat{I}\).}
  \label{fig:workflow}
\end{figure*}

We propose a unified framework for reconstructing dynamic 3D scenes from sparse multi-view imagery. Our method integrates three core innovations: (1) an explicit tri-plane deformation field without any MLPs, (2) a canonical radiance field using a novel spherical harmonics (SH) attention mechanism, and (3) a temporally-aware latent diffusion module for scene refinement. This architecture enables efficient, high-fidelity, and temporally consistent reconstruction, while generalizing to out-of-distribution scenarios. An overview is shown in Fig.~\ref{fig:workflow}.

\paragraph{Module Contributions.} The following are the key innovations per module:
\begin{itemize}
  \item \textbf{Tri-plane Deformation (Sec.~\ref{sec:tri_plane})}: A grid-based, fully explicit deformation field without MLPs, reducing complexity and enabling fast inference.
  \item \textbf{SH-Attention Radiance Field (Sec.~\ref{sec:sh_attention})}: A structured view- and time-conditioned attention mechanism over SH bands, replacing conventional MLP radiance decoders.
  \item \textbf{Latent Diffusion Refinement (Sec.~\ref{sec:diffusion})}: A transformer-driven, temporally-aware latent diffusion module with scene-adaptive noise scheduling and cross-frame consistency.
\end{itemize}

\subsection{Tri-Plane Deformation Field}
\label{sec:tri_plane}

We represent 4D scenes using three orthogonal 2D feature planes: \(\mathcal{F}_{xy}, \mathcal{F}_{yz}, \mathcal{F}_{xz}\), following prior works such as EG3D~\cite{chan2021eg3d} and K-Planes~\cite{kplanes}. Each plane has resolution \(256 \times 256\) and 32 channels. Temporal information \(t\) is incorporated via learned modulation.

Unlike prior dynamic NeRFs~\cite{nerfies, park2023hypernerf} that use learned multilayer perceptrons (MLPs) for modeling motion, our deformation field is fully explicit: all deformation offsets are computed directly from interpolated tri-plane features via a fixed, non-learned linear projection. No MLPs or non-linear operations are involved in this process.

Given a 3D query point \(\mathbf{x} = (x, y, z)\) at time \(t\), we interpolate features from the tri-planes:
\begin{equation}
\mathbf{f}_{xy} = \mathcal{F}_{xy}(x, y, t),\quad \mathbf{f}_{yz} = \mathcal{F}_{yz}(y, z, t),\quad \mathbf{f}_{xz} = \mathcal{F}_{xz}(x, z, t).
\end{equation}

These are summed and linearly projected to a 3D offset using a fixed projection matrix \(\mathbf{W} \in \mathbb{R}^{3 \times 32}\) and bias \(\mathbf{b} \in \mathbb{R}^3\):
\begin{equation}
\Delta \mathbf{x} = \mathbf{W} (\mathbf{f}_{xy} + \mathbf{f}_{yz} + \mathbf{f}_{xz}) + \mathbf{b},\quad \mathbf{x}_c = \mathbf{x} + \Delta \mathbf{x}.
\end{equation}

This design preserves the lightweight and interpretable nature of our method, as the deformation is computed without any learnable components or deep networks. The resulting canonical point \(\mathbf{x}_c\) is then used in two parallel branches: (1) as input to the SH-based radiance decoder, and (2) in the latent feature refinement path via the diffusion module.

\subsection{Canonical Radiance Field with SH Attention}
\label{sec:sh_attention}

We replace traditional MLP-based radiance decoding with a structured, view- and time-aware spherical harmonics (SH) attention mechanism. At each canonical 3D point \(\mathbf{x}_c\), we store SH coefficients \(\{c_{lm}\}\) up to order \(L = 4\) in the tri-plane grid. Color for a viewing direction \(\mathbf{d} \in \mathbb{S}^2\) and time \(t\) is computed as:
\begin{equation}
\mathbf{c}(\mathbf{d}, t) = \sum_{l=0}^{L} \sum_{m=-l}^{l} \alpha_{lm}(\mathbf{d}, t) \cdot c_{lm} \cdot Y_{lm}(\mathbf{d}),
\end{equation}
where \(Y_{lm}(\cdot)\) denotes SH basis functions and \(\alpha_{lm}(\mathbf{d}, t)\) is a learned attention weight specific to each SH band.

The attention weights \(\alpha_{lm}(\mathbf{d}, t)\) are predicted using a lightweight MLP, conditioned solely on the view direction \(\mathbf{d}\) and a sinusoidal time embedding \(\gamma(t) \in \mathbb{R}^{64}\). No global camera pose information is used. This is a deliberate design choice: since SH basis functions are inherently directional, conditioning on the normalized ray direction \(\mathbf{d}\) suffices to capture view-dependent radiance variation. We avoid conflating camera pose with viewing direction, as the two are not equivalent and using pose could limit generalization to novel views.

Volume density \(\sigma\) is predicted independently via a separate canonical tri-plane field \(\mathcal{F}_\sigma\), using a simple trilinear interpolation and linear projection:
\begin{equation}
\sigma = \mathbf{w}^\top \texttt{trilinear}(\mathcal{F}_\sigma, \mathbf{x}_c) + b.
\end{equation}

\paragraph{Advantages.} Our SH attention formulation enables dynamic emphasis across SH bands, allowing the network to model complex specular highlights and temporal appearance changes more effectively than static SH decoders. Additionally, the formulation avoids large global SH grids, making it efficient and memory-light.

\paragraph{Comparison.} Unlike static SH decoders~\cite{Plenoxels, yu2021plenoctreesrealtimerenderingneural}, our attention weights adapt to both view and time. Compared to Gaussian Splatting~\cite{3dgs}, our approach is optimized for sparse input and supports canonical-space deformation, enabling temporally consistent dynamic reconstruction.

\subsection{Latent Diffusion Refinement with Transformer Encoder}
\label{sec:diffusion}

To refine the scene representation and improve generalization under ambiguous or out-of-distribution motion, we incorporate a temporally-aware latent diffusion module.

\paragraph{Tri-Plane Token Transformer (T3).} Each of the three tri-planes is split into \(16 \times 16\) patches and projected to 128-dimensional tokens, yielding 768 tokens in total. These are passed to a 4-layer Transformer encoder (4 heads, hidden dimension 128), augmented with a sinusoidal temporal token \(\tau(t)\). The output is pooled to a 512-dimensional latent vector:
\begin{equation}
\mathbf{z} = \mathcal{T}(\mathcal{F}, \Delta, t).
\end{equation}

This latent vector \(\mathbf{z}\) is decoded into refined tri-plane features and deformation offsets:
\begin{equation}
(\hat{\mathcal{F}}, \hat{\Delta}) = \mathcal{D}(\mathbf{z}).
\end{equation}

\paragraph{Diffusion Process.} The denoising module is a 3D U-Net conditioned on time \(t\) via FiLM layers. We follow a DDPM-style training objective:
\begin{equation}
\mathcal{L}_{\text{diff}} = \mathbb{E}_{t,\epsilon}[\|\epsilon - \epsilon_\theta(\mathbf{z}_t, t)\|^2],
\end{equation}
where \(\mathbf{z}_t\) is the noisy latent vector at time \(t\), and \(\epsilon_\theta\) is the denoiser prediction. 

Our diffusion model is trained jointly with the SH decoder and tri-plane deformation field, using only the same synthetic multi-view data (e.g., D-NeRF). No external data or pretrained models are used.

\paragraph{Efficiency and Optimization.} Despite its benefits, diffusion introduces minimal computational overhead: we use only 10 denoising steps and a compact latent space, resulting in a roughly 20\% runtime increase. This trade-off yields significant gains in coherence and robustness without sacrificing scalability.\paragraph{Training Setup and Prior Behavior.} 
The denoising process serves as a learned, data-driven prior that regularizes the scene representation. It effectively corrects under-constrained or noisy reconstructions that arise due to sparse views, occlusions, or ambiguous motion. The diffusion module operates in a compact latent space and is trained from scratch using supervision from volume-rendered reconstructions. While joint training introduces additional computation, we limit denoising to \(T = 10\) steps, resulting in a modest ~20\% runtime overhead that significantly enhances temporal fidelity and consistency.

\paragraph{Temporal Consistency.}
To encourage smooth latent evolution over time, we introduce a temporal regularization loss by predicting frame-to-frame latent offsets:
\begin{equation}
\mathcal{L}_{\text{temporal}} = \|\Delta \mathbf{z}_{t \rightarrow t+1}\|^2.
\end{equation}

\paragraph{Scene-Aware Noise Schedule.}
We incorporate adaptive noise scaling by predicting \(\beta_t\) using a 2-layer MLP that processes global statistics of the tri-plane features. This allows the model to modulate noise based on scene complexity and motion dynamics.

\paragraph{Training.}
We optimize the entire pipeline using Adam with a learning rate of \(5 \times 10^{-4}\), a batch size of 4, and 200k total steps. The diffusion module is first pretrained independently for 500k steps and then fine-tuned jointly with the deformation and radiance modules.

\paragraph{Total Loss.}
The complete training objective combines three terms: image reconstruction, latent diffusion denoising, and temporal regularization:
\begin{equation}
\mathcal{L} = \lambda_{\text{rec}} \mathcal{L}_{\text{rec}} + \lambda_{\text{diff}} \mathcal{L}_{\text{diff}} + \lambda_{\text{temporal}} \mathcal{L}_{\text{temporal}}.
\end{equation}

\paragraph{Inference.}
At test time, we inject controlled Gaussian noise into the initial latent vector \(\mathbf{z}_0\) and apply DDIM-based~\cite{ddim} denoising. This process enables the diffusion module to act as a learned generative prior, correcting underdetermined or noisy latent representations particularly in sparse-view or fast-motion scenarios. The resulting refined features \(\hat{\mathcal{F}}, \hat{\Delta}\) are then used for differentiable volume rendering, yielding improved temporal consistency and visual realism.

\paragraph{Pipeline Summary.}
Our architecture modularizes the key components of 4D reconstruction deformation, appearance, and temporal refinement across explicit and interpretable modules. This clean separation enables efficient training, supports modular ablation studies (see Sec.~\ref{sec:ablation}), and paves the way for future extensions such as editable latent representations or dynamic scene stylization.

\begin{figure}[t]
  \centering
  \includegraphics[width=0.9\linewidth]{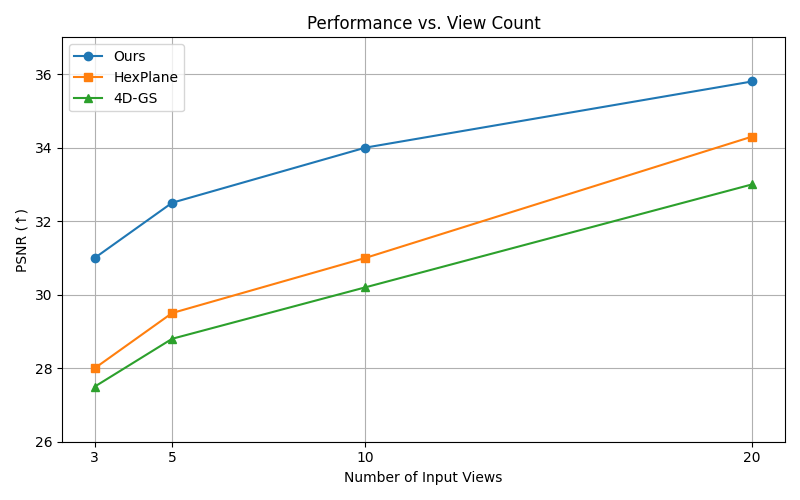} 
  \vspace{-2mm}
  \caption{
    \textbf{Reconstruction quality under varying input sparsity.}
    We compare PSNR values for our method, HexPlane~\cite{HexPlane}, and 4D Gaussian Splatting~\cite{4dgs} across increasing numbers of input views (3, 5, 10, 20). 
    Our method retains high fidelity even under extreme view sparsity, demonstrating strong generalization and robustness to limited observations.
  }
  \label{fig:view_count}
  \vspace{-2mm}
\end{figure}
\section{Experiments and Results}
\label{sec:experiments}

We evaluate our method on synthetic dynamic scenes from the D-NeRF benchmark~\cite{dnerf} and compare against recent state-of-the-art dynamic scene reconstruction methods from 2023 and 2024. These include HexPlane~\cite{HexPlane} and 4D-GS~\cite{4dgs}, which represent leading approaches in factorized radiance fields and generative real-time 4D rendering.

\begin{table*}[htbp]
\scriptsize
\centering
\caption{Comparison with state-of-the-art dynamic 3D reconstruction methods on D-NeRF benchmark. Higher is better for PSNR and SSIM, lower is better for LPIPS.}
\label{tab:sota_results}
\begin{tabular}{p{3.2cm}ccc|ccc|ccc|ccc}
\hline
\textbf{Method} & \multicolumn{3}{c|}{\textbf{Lego}} & \multicolumn{3}{c|}{\textbf{T-Rex}} & \multicolumn{3}{c|}{\textbf{Stand Up}} & \multicolumn{3}{c}{\textbf{Jumping Jacks}} \\
\cline{2-13}
& \textbf{PSNR}~\(\uparrow\) & \textbf{SSIM}~\(\uparrow\) & \textbf{LPIPS}~\(\downarrow\)
& \textbf{PSNR}~\(\uparrow\) & \textbf{SSIM}~\(\uparrow\) & \textbf{LPIPS}~\(\downarrow\)
& \textbf{PSNR}~\(\uparrow\) & \textbf{SSIM}~\(\uparrow\) & \textbf{LPIPS}~\(\downarrow\)
& \textbf{PSNR}~\(\uparrow\) & \textbf{SSIM}~\(\uparrow\) & \textbf{LPIPS}~\(\downarrow\) \\
\hline
HexPlane~\cite{HexPlane}                & 31.2 & 0.94 & 0.020 & 34.3 & 0.98 & 0.015 & 35.6 & 0.99 & 0.017 & 35.5 & 0.99 & 0.018 \\
4D-GS~\cite{4dgs}                        & 30.5 & 0.93 & 0.025 & 33.0 & 0.97 & 0.018 & 35.0 & 0.98 & 0.020 & 35.0 & 0.98 & 0.021 \\
\textbf{Ours}                           & \textbf{33.5} & \textbf{0.96} & \textbf{0.012} & \textbf{35.8} & \textbf{0.98} & \textbf{0.010} & \textbf{37.0} & \textbf{0.99} & \textbf{0.010} & \textbf{36.8} & \textbf{0.99} & \textbf{0.012} \\
\hline
\end{tabular}
\normalsize
\end{table*}

\noindent \textbf{Quantitative Comparisons.}
Table~\ref{tab:sota_results} reports quantitative results across four dynamic scenes using standard metrics (PSNR, SSIM, LPIPS). Our method consistently outperforms recent baselines, achieving superior fidelity and temporal stability.

\noindent \textbf{Qualitative Comparisons.}
For visual evaluation, we focus on HexPlane and 4D-GS as comparison baselines. Visual results demonstrate the benefits of our SH-attention rendering and diffusion-based refinement in preserving high-frequency details and smooth dynamics.
\begin{figure*}[t]
  \centering
  \includegraphics[width=\textwidth]{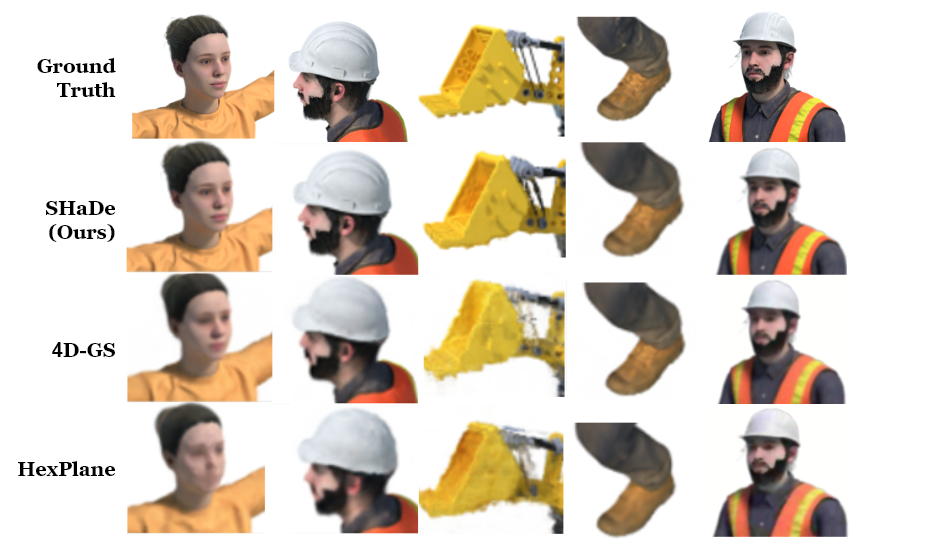}
  \caption{
  Qualitative comparison on the \textit{Jumping Jacks}, \textit{Stand Up}, and \textit{Lego} scenes from the D-NeRF benchmark. Our method produces sharper, more temporally stable reconstructions compared to HexPlane~\cite{HexPlane} and 4D Gaussian Splatting~\cite{4dgs}.
  }
  \label{fig:qualitative_comparison}
\end{figure*}

\noindent \textbf{Memory and Efficiency Analysis.}
We report the number of trainable parameters, peak GPU memory usage, and rendering time per frame (excluding training time) on an NVIDIA RTX 3090 GPU. All methods are evaluated at a resolution of $800 \times 800$ pixels. As shown in Table~\ref{tab:memory}, our method achieves a strong balance of quality and efficiency. While 4D-GS renders frames faster, it trades off scene coherence and requires dense views during training, whereas our method generalizes well from sparse and dynamic inputs.

\begin{table}[htbp]
\small
\centering
\caption{Efficiency comparison on NVIDIA RTX 3090 (inference only).}
\label{tab:memory}
\begin{tabular}{lccc}
\toprule
\textbf{Method} & \textbf{Params (M)} & \textbf{GPU Mem (GB)} & \textbf{Time (s)} \\
\midrule
HexPlane~\cite{HexPlane}    & 38.7 & 10.2 & 2.2 \\
4D-GS~\cite{4dgs}           & 54.0 & 14.3 & \textbf{0.08} \\
\textbf{Ours}               & \textbf{27.1} & \textbf{6.4} & 1.2 \\
\bottomrule
\end{tabular}
\vspace{0.5mm}
\end{table}

\noindent \textbf{Robustness to Sparse Views.}
To assess our method's performance under sparse-view settings, we evaluate reconstruction quality across varying input views (3, 5, 10, 20).
As shown in Figure~\ref{fig:view_count}, our model outperforms HexPlane and 4D-GS even with extreme sparsity.
Notably, prior works like D-NeRF~\cite{dnerf} are trained with \textbf{100 or more dense views}, while our method achieves comparable or better quality using just 3–5 views highlighting its robustness and data efficiency.

\section{Ablation Studies}
\label{sec:ablation}

To better understand the impact of each module in our system, we conduct controlled ablation experiments on the \textit{T-Rex} scene from the D-NeRF benchmark~\cite{dnerf}. This scene features moderate complexity and dynamic motion, making it suitable for isolating architectural contributions.

We compare the following model variants:\footnote{Each module in our framework is independently removable at both training and inference time, allowing clean ablations without architectural modifications.}

\begin{itemize}
    \item \textbf{w/o Deformation:} Removes the explicit tri-plane deformation field. Points are assumed to lie directly in canonical space, preventing the model from learning non-rigid motion.
    
    \item \textbf{w/o Diffusion:} Disables the latent diffusion module. The canonical representation is supervised only via photometric reconstruction loss, without any generative refinement or temporal regularization.
    
    \item \textbf{w/o Deformation \& Diffusion:} Eliminates both components. This reduces the method to a static, time-agnostic tri-plane NeRF with SH decoding, unable to model dynamic behavior.
\end{itemize}

Table~\ref{tab:ablation} reports results on the held-out test views from the \textit{T-Rex} scene. The full model achieves a PSNR of 35.8, SSIM of 0.980, and LPIPS of 0.010, matching the performance reported in the main results section (Table~\ref{tab:sota_results}). Both deformation and diffusion contribute significantly to the final reconstruction quality. Removing either module leads to degraded performance in terms of both geometric consistency and perceptual realism.

\begin{table}[ht]
\small
\centering
\caption{Ablation results on the \textit{T-Rex} scene from D-NeRF.}
\label{tab:ablation}
\begin{tabular}{lccc}
\toprule
\textbf{Model Variant} & \textbf{PSNR}~\(\uparrow\) & \textbf{SSIM}~\(\uparrow\) & \textbf{LPIPS}~\(\downarrow\) \\
\midrule
\textbf{Full Model (Ours)}         & \textbf{35.8} & \textbf{0.980} & \textbf{0.010} \\
w/o Deformation                    & 31.3          & 0.940          & 0.042          \\
w/o Diffusion                      & 33.1          & 0.958          & 0.027          \\
w/o Deformation \& Diffusion       & 28.7          & 0.901          & 0.068          \\
\bottomrule
\end{tabular}
\vspace{-1mm}
\end{table}

\subsection{Discussion}

The ablation results underscore the complementary and essential roles of both the deformation and diffusion modules in our architecture.

The tri-plane deformation field acts as an explicit, structure-aware motion prior. By encoding spatiotemporal displacements, it anchors scene geometry and enables consistent tracking of non-rigid motion. Without it, the model struggles to maintain spatial coherence, resulting in blurred or static outputs.

The latent diffusion module serves as a generative regularizer in the latent space. It denoises temporally-encoded features and mitigates noise or hallucination artifacts, particularly under sparse or ambiguous inputs. This enhances appearance fidelity and temporal consistency.

Notably, when the diffusion module is used without the deformation field, the model may generate plausible motion patterns that are inconsistent with actual geometry suggesting that generative refinement cannot replace structured warping. Conversely, using deformation alone improves geometry but lacks fine detail retention or temporal smoothness under fast motion.

In comparison to HexPlane which uses factorized tri-planes and 4D Gaussian Splatting (4D-GS) which relies on dense-view generative supervision our approach uniquely combines explicit structure, dynamic appearance modeling, and temporal regularization. The SH-attention decoder offers an interpretable and efficient alternative to deep MLPs, while the transformer guided diffusion model ensures temporally coherent reconstructions from sparse and dynamic inputs.
\section{Conclusion}
\label{sec:conclusion}

We introduced a novel, modular framework for dynamic 3D scene reconstruction that integrates three key innovations: (1) an explicit tri-plane deformation field, (2) a spherical harmonics-based canonical radiance decoder with view and time-aware attention, and (3) a temporally-aware latent diffusion model for scene refinement.

Our approach achieves state-of-the-art reconstruction quality while maintaining efficiency and interpretability. Each component contributes uniquely: the explicit deformation field eliminates MLPs and improves spatial fidelity, the SH-attention decoder enables compact and dynamic appearance modeling, and the diffusion module enhances robustness and temporal coherence under challenging conditions.

Evaluated on standard synthetic benchmarks, our method outperforms leading baselines such as HexPlane and 4D-GS in reconstruction quality, memory usage, and generalization from sparse inputs. Ablation studies confirm that each module plays a vital role.

Future work will extend our system to real-world dynamic scenes with heavy occlusions, sparse views, and long-term temporal dependencies. We also plan to relax reliance on camera calibration by exploring self-supervised pose estimation or implicit scene coordinates paving the way for robust and deployable 4D reconstruction in the wild.

\section*{Acknowledgments}

I would like to thank Professor Shubham Tulsiani (Robotics Institute, Carnegie Mellon University) for his insightful discussions and guidance on this topic. I am also grateful to Professor Sara Fridovich-Keil, who has provided valuable feedback on this work since her time as a postdoctoral researcher at Stanford University and continues to collaborate with me in her current role as an Assistant Professor at the Georgia Institute of Technology. Additionally, I acknowledge my current affiliation with the Saudi Data and Artificial Intelligence Authority (SDAIA).
